# New Arabic Medical Dataset for Diseases Classification

Jaafar Hammoud[1], Aleksandra Vatian[1], Natalia Dobrenko[1], Nikolai Vedernikov[1], Anatoly Shalyto[1] and Natalia Gusarova[1]

[1] ITMO University, Kronverksky pr. 49, Saint Petersburg, Russia
`hammoudgj@gmail.com`

**Abstract.** The Arabic language suffers from a great shortage of datasets suitable for training deep learning models, and the existing ones include general non-specialized classifications.
In this work, we introduce a new Arab medical dataset, which includes two thousand medical documents collected from several Arabic medical websites, in addition to the Arab Medical Encyclopedia. The dataset was built for the task of classifying texts and includes 10 classes (Blood, Bone, Cardiovascular, Ear, Endocrine, Eye, Gastrointestinal, Immune, Liver and Nephrological) diseases.
Experiments on the dataset were performed by fine-tuning three pre-trained models: BERT from Google, Arabert that based on BERT with large Arabic corpus, and AraBioNER that based on Arabert with Arabic medical corpus.

## 1 Introduction

These days, unstructured text is everywhere, from our conversations and comments on social media to emails, websites, etc. Their processing by means of artificial intelligence (AI) techniques and Natural Language Processing (NLP) is at a high level [1], including such an important type of text as medical records. Extracting useful information from medical texts and reports automatically plays a pivotal and important role in supporting medical decision making [2, 3, 4].

Unfortunately, AI and NLP tools are used to a much lesser extent for processing texts in Arabic, especially in the medical field. This can be attributed to several reasons, the most important of which is that the use of electronic medical reports and records was not common in most Arabic countries. The handwritten and not digitally structured reports are still in common use in the majority of Arab countries. In addition, the Arab countries, which are somehow more advanced than their counterparts, use English as the main language in their medical records. Therefore, the Arabic language faces a clear shortage in the availability of appropriate data to train models capable of contributing to improving the quality of health care and helping doctors diagnose and treat [5].

In this work, we present a new Arabic dataset specialized in classifying medical texts and show its use on three NLP models.

## 2   Related Work

In the period extending from the sixties of the last century until the end of the first decade of the current one, shallow classification algorithms were dominant, which depend on the process of extracting properties and features manually, which requires hiring experts in the field to deal with the data that the algorithms learn on.

Shallow learning algorithms are affected by the preprocessing process a lot, from the segmentation of words, data cleaning, to conducting an accurate statistical study about it, and then the representation of data in a way that is easier for computers, such as the Bag-of-words (BOW) [6], GloVe [7], word2vec [8], N-gram [9] and TF-IDF [10]. These methods generate a vector representing properties extracted from the text that will then feed the classifier.

The Bayesian network [11], and the hidden Markov network [12], is a combination of probability theory and graph theory. It is considered as a probabilistic graphical model that expresses conditional dependencies between features in graphs.

Naïve-Bayes (NB) [13] is one of the simplest and commonly used models for classification. Usually, the NB algorithm is mainly based on the use of the former probability to calculate the later probability.

The hidden Markov model is suitable for dealing with data in the form of series such as texts and is considered one of the effective models in reducing computational complexity by redesigning the structure of the model, where researchers [14] used this model to implement the task of classifying medical text.

K-Nearest Neighbors algorithm [15] is used to classify unlabeled data and that is done without building the model, tune several parameters, or make additional assumptions. In [16] simple classification algorithms appropriate for annotating textual materials with partial information have been presented. As a result, they are suitable for large multi-label classification, especially in the biomedical domain.

Where the SVM method [17] turns text classification tasks into multiple binary classification tasks, the researchers [18] used Machine-generated regular expressions effectively for clinical text classification. They combined a regular expression-based classifier with SVM, to improve classification performance.

The researchers [19] proposed a HDTTCA approach to identify those who are eligible for telehealth services, which is a systematic approach that involves data set preprocessing, decision tree model building, and predicting and explaining the most important attributes in the data set for patients.

After 2010 the usage of deep learning models started to grow up, some research papers discussed models based on RNN and Long Short-Term Memory (LSTM) [20], Tree-LSTM [21], Multi-Timescale [22], Bidirectional-LSTM with two-dimensional max-pooling [23], some papers used the impact of CNN architectures with word embeddings, researchers [24] presented a Very Deep CNN model for text processing inspired by VGG [25] and ResNets [26]. One of other interesting CNN-based models [27] presented a tree-based CNN to capture sentence level semantics. There is also a growing interest in applying CNNs to biomedical text classification [28–31].

Both RNN and CNN models suffer from dealing with long sequential, and the computational cost increases with the increase of the length of the sentence. Transformers [32] came to address this shortcoming, by applying self-attention by calculating the "attention score" to the impact of each word on the other in a parallel way. since 2018 a set of large-scale Transformer-based Pre-trained Language Model was launched [20] and they achieve today state-of-the-art results in NLP tasks, like BERT [33], BERT with nonlinear gradient method [34], openGPT [35, 36], RoBERTa [37], and ALBERT [38].

## 3  Dataset

The Arabic language suffers from a scarcity in the availability of appropriate datasets for training. There are datasets available for the task of classification, but it is not specialized in a particular field, but rather it is data collected from popular news websites in the Arab world, for example, the KACST [39] corpus was collected from the Saudi Press Agency, Arabic poems, and discussion forums, the BBC corpus[1] was collected from BBC Arabic website. This corpus has 4763 documents, with 7 categories, the CNN corpus[2] was collected from CNN Arabic website. This corpus has 5070 documents, with 6 categories and The OSAC [40] was collected with a crawler from dozens of Arabic websites. All these datasets provide records for classifying topics within general classes (such as politics, sports, economics, etc.)

In this work, we present a new Arabic dataset specialized in classifying medical texts. This dataset was collected semi-automatically by several libraries available in the Python language (Request, Beautiful Soup, and Selenium) from several Arab medical websites and from the Arabic Medical Encyclopedia. In its first version, the dataset includes 2,000 medical documents divided into 10 categories, the following Table 1 is a detailed description of these dataset.

**Table 1**. Summary of the dataset

| Sources | Class | N | W | S |
|---|---|---|---|---|
| altibbi.com | Blood disease | 215 | 1251.7 | 25.3 |
|  | Bone diseases | 211 | 1325.3 | 26.8 |
| webteb.com | Cardiovascular diseases | 195 | 1749.5 | 27.1 |
|  | Ear diseases | 180 | 1307.5 | 23.9 |
| mayoclinic.org | Endocrine diseases | 204 | 1184.6 | 22.4 |
|  | Eye diseases | 190 | 1456.1 | 26.8 |
| dailymedi-calinfo.com | Gastrointestinal diseases | 218 | 1381.6 | 25.9 |
|  | Immune diseases | 203 | 1253.2 | 24.1 |
|  | Liver diseases | 198 | 1386.7 | 27.3 |
| arab-ency.com.sy/medi-cal/ | Nephrological diseases | 186 | 1078.1 | 22.9 |

---

[1] https://sourceforge.net/projects/newarabiccorpus/
[2] https://osdn.net/projects/sfnet_ar-text-mining/downloads/Arabic-Corpora/cnn-arabic-utf8.7z/

Where N denotes the number of samples, and W and S the average number of words and sentences per document, respectively.

The process of verifying the correct classification of each document was carried out by three doctors working in Syrian university hospitals. This sample of dataset is part of a project that aims to provide a medical Arabic dataset that supports the following tasks (Text Classification, Named Entity Recognition, Question-answering system), by the end of this project and when providing this dataset to the public, it will be available without any preprocessing operations on the text, that allows researchers to implement operations that serve their purposes, in addition, the pre-trained models having their own preprocessing operations (for example, removing stop words may have a negative impact on training accuracy), as these models are able to take advantage of the full context of a sentence.

## 4   Methodology and Experiments

To classify our medical documents presented in Table 1 into ten categories (Blood, Bone, Cardiovascular, Ear, Endocrine, Eye, Gastrointestinal, Immune, Liver, and Nephrological) diseases, three pre-trained models (BERT, AraBERT, BioAraBert) that based on Transformers and two shallow algorithms (SVM, Naive Bayes) were used.

### 4.1   BERT

BERT [33] employs a transformer that consists of two distinct mechanisms (encoder and decoder), with the encoder reading the text input and the decoder producing a task prediction. Only the encoder is required because the BERT's purpose is to construct a language model. The Transformer encoder is called bidirectional since it reads the full sequence of words at once, rather than (left-to-right or right-to-left).

BERT has two training techniques. The first is known as "Masked LM," in which the model substitutes 15% of the words in each sequence with a [MASK] token and attempts to predict the masked words' original value. The second technique is called "Next Sentence Prediction (NSP)" and it involves the model receiving pairs of phrases and attempting to predict whether the second sentence in the pair is the same as the second sentence in the original document.

### 4.2   Arabert

While BERT has been trained on 3.3 billion words extracted from the English Wikipedia and the Book Corpus [41], the Arabic Wikipedia is small compared to its English counterpart. The researchers in [42] scraped manually articles from Arabic news websites, and they used two large Arabic corpus (1.5 billion words Arabic [43], and OSIAN corpus [44]).

With using all these corpuses, the final size for pre-training dataset is 70 million sentences (~24 GB) of text.

4### 4.3 ABioNER

The pretraining ABioNER [45] used the AraBERTv0.1-base in addition to Arabic biomedical literature corpus which is collected from (PubMed, MedlinePlus Health Information in Arabic[3], Journal of Medical and Pharmaceutical Sciences[4], Arab Journal for Scientific Publishing[5], Eastern Mediterranean Health Journal [46]). Fig 2 Shows AraBioNER structure.

**Fine-Tuning**
The dataset was divided into 80% for training, 10% for validation and 10% for testing. For fine-tuning we used (BERT-Base, Multilingual Cased model, Arabert v2, ABioNER that is based on Arabert v1).

For all three models we used Adam optimizer with learning rate $lr = 1e - 4$ and with two fully connected dense layers of size 1024 and 10 respectively, first one with relu activation function, and second one with softmax activation function, all that done for 4 epochs and 16 batch size.

We measure the effectiveness of the model using the traditional metrics:

$$F1 = 2 * \frac{Precision * Recall}{Precision + Recall}, Precision = \frac{TP}{TP + FP}, Recall = \frac{TP}{TP + FN}$$

Here TP denotes to true positive, FP to false positive, and FN to false negative.

While fine-tuning the three models, we optimized the batch size, maximum sequence length (MSL), number of epochs, number of tokens that documents are truncated to, and learning rate.

For our dataset, we find that using a batch size of 16, a learning rate of (2e-5), and an MSL of 512 tokens provides the best results. As is the case with [33, 47].

To compare with shallow algorithms, we took the most popular and efficient ones in literature SVM and Naïve Bayes. We mentioned in the introduction that shallow algorithms are greatly affected by the preprocessing operations, Therefore, we applied to our medical text a set of preprocessing operations, first tokenization by NLTK[6] library, then removing stop words, and then stemming by Snowball[7] stemmer, and finally, we selected TF-IDF that we mentioned in the Introduction too for word vectorization. The training for shallow algorithms was carried out using a scikit-learn library, the results came as shown in the Table 2.

**Table 2.** F1 score for the three models

|  | BERT | Arabert v2 | ABioNER | SVM | NB |
|---|---|---|---|---|---|
| F1. Validation | 94.1343 | 96.4327 | **97.4331** | 89.1308 | 87.6118 |
| F1. Testing | 92.2934 | 94.5415 | **95.9124** | 87.3473 | 85.6949 |

---

[3] https://medlineplus.gov/languages/arabic.html
[4] https://www.ajsrp.com/journal/index.php/jmps
[5] https://www.ajsp.net/
[6] https://www.nltk.org/
[7] https://snowballstem.org/

The results show that the three models BERT, Arabert and ABioNER are more accurate and efficient than the shallow algorithms, and this is not inconsistent with the literature. The use of pre-trained models on a huge amount of data and fine-tuning them on a specific dataset is state-of-the-art these days.

From Table 2, the ABioNER model achieved the best accuracy, because this model is pre-trained on the same multilingual BERT's dataset and Arabert's Arabic dataset in addition to a large corpus of a medical dataset in Arabic.

For shallow algorithms, the good results shown in Table 2 show that the dataset that we have presented is balanced and suitable as a benchmark for future work aimed at dealing with Arabic medical texts.

**Conclusion**

In this work, we presented a new Arabic medical dataset, for the text classification task, the dataset includes 2000 articles, with 10 classes (Blood, Bone, Cardiovascular, Ear, Endocrine, Eye, Gastrointestinal, Immune, Liver, and Nephrological) diseases.

Through experiments, we have found that pre-trained models that have trained on large related corpus, and fine-tuned with specific datasets yield state-of-the-art results. This is evident by comparing the ABioNER model that has pre-trained on Arabic medical corpus before fine-tuned on our dataset with the original BERT model.

8